\documentclass[journal]{IEEEtran}
\usepackage{amsmath, amssymb}
\usepackage{graphicx}
\usepackage{comment}
\usepackage{multirow}
\usepackage{booktabs}
\usepackage{textcomp}
\usepackage{float}
\usepackage{enumitem}
\usepackage{bm}
\restylefloat{table}
\usepackage{subfigure}
\usepackage{caption}
\usepackage{ctable}
\usepackage{colortbl}
\definecolor{Gray}{gray}{0.9}
\newtheorem{definition}{Definition}[section]

\begin{document}
	
%\title{Attention Models with Random Features for Multi-layered Graph Embeddings}
\title{GrAMME: Semi-Supervised Learning using Multi-layered Graph Attention Models}

\author{Uday Shankar Shanthamallu$^*$,
        Jayaraman~J.~Thiagarajan$^\dagger$,
        Huan~Song$^\ddagger$,
        and~Andreas~Spanias$^*$,~\IEEEmembership{Fellow,~IEEE}
\thanks{$^*$SenSIP Center, School of ECEE, Arizona State University, $^\dagger$Lawrence Livermore National Labs, $^\ddagger$Bosch Research North America}}

\markboth{GrAMME: Semi-Supervised Learning using Multi-layered Graph Attention Models}%
{GrAMME: Semi-Supervised Learning using Multi-layered Graph Attention Models}

\maketitle

\begin{abstract}
Modern data analysis pipelines are becoming increasingly complex due to the presence of multi-view information sources. While graphs are effective in modeling complex relationships, in many scenarios a single graph is rarely sufficient to succinctly represent all interactions, and hence multi-layered graphs have become popular. Though this leads to richer representations, extending solutions from the single-graph case is not straightforward. Consequently, there is a strong need for novel solutions to solve classical problems, such as node classification, in the multi-layered case. In this paper, we consider the problem of semi-supervised learning with multi-layered graphs. Though deep network embeddings, e.g. DeepWalk, are widely adopted for community discovery, we argue that feature learning with random node attributes, using graph neural networks, can be more effective. To this end, we propose to use attention models for effective feature learning, and develop two novel architectures, GrAMME-SG and GrAMME-Fusion, that exploit the inter-layer dependencies for building multi-layered graph embeddings. Using empirical studies on several benchmark datasets, we evaluate the proposed approaches and demonstrate significant performance improvements in comparison to state-of-the-art network embedding strategies. The results also show that using simple random features is an effective choice, even in cases where explicit node attributes are not available.
\end{abstract}


% \fi
\IEEEpeerreviewmaketitle

\section{Introduction}
\label{sec:intro}
\subsection{Multi-layered Graph Embeddings}
The prevalence of relational data in several real-world applications, e.g. social network analysis \cite{eagle2006reality}, recommendation systems \cite{rao2015collaborative} and neurological modeling \cite{fornito2013graph}, has led to crucial advances in machine learning techniques for graph-structured data. This encompasses a wide-range of formulations to mine and gather insights from complex network datasets -- node classification~\cite{kipf2016semi}, link prediction~\cite{zhang2018}, community detection~\cite{newman2006finding}, influential node selection \cite{li2017influential} and many others. Despite the variabilities in these formulations, a recurring idea that appears in almost all of these approaches is to obtain 
embeddings for nodes in a graph, prior to carrying out the downstream learning task. In the simplest form, the adjacency matrix indicating the connectivities can be treated as na\"ive embeddings for the nodes. However, it is well known that such \textit{cursed}, high-dimensional representations can be ineffective for the subsequent learning. Hence, there has been a long-standing interest in constructing low-dimensional embeddings that can best represent the network topology. 

Until recently, the majority of existing work has focused on analysis and inferencing from a single network. However, with the emergence of multi-view datasets in real-world scenarios, commonly represented as \textit{multi-layered} graphs, conventional inferencing tasks have become more challenging. Our definition of multi-layered graphs assumes complementary views of connectivity patterns for the same set of nodes, thus requiring the need to model complex dependency structure across the views. The heterogeneity in the relationships, while providing richer information, makes statistical inferencing challenging. Note that, alternative definitions for multi-view networks exist in the literature~\cite{li2018multi}, wherein the node sets can be different across layers (e.g. interdependent networks). Prior work on multi-layered graphs focuses extensively on unsupervised community detection, and they can be broadly classified into methods that obtain a consensus community structure for producing node embeddings \cite{dong2012clustering,dong2014clustering,kim2017differential,tagarelli2017ensemble}, and methods that infer a separate embedding for a node in every layer, while exploiting the inter-layer dependencies, and produce multiple potential community associations for each node \cite{mucha2010community,bazzi2016community}. In contrast to existing approaches, the goal of this work is to build multi-layered graph embeddings that can lead to improved node label prediction in a semi-supervised setting.

\subsection{Deep Learning on Graphs}
Node embeddings can be inferred by optimizing with a variety of measures that describe the graph structure -- examples include decomposition of the graph Laplacian \cite{ng2002spectral}, stochastic factorization of the adjacency matrix \cite{ahmed2013distributed}, decomposition of the modularity matrix \cite{newman2006finding,chen2014community} etc. The unprecedented success of deep learning with data defined on regular domains, e.g. images and speech, has motivated its extension to arbitrarily structured graphs. For example, Yang \textit{et al.}~\cite{yang2016modularity} and Thiagarajan \textit{et al.}~\cite{thiagarajan2016robust} have proposed stacked auto-encoder style solutions, that directly transform the objective measure into an undercomplete representation. An alternate class of approaches utilize the distributional hypothesis, popularly adopted in language modeling \cite{harris1954distributional}, where co-occurrence of two nodes in short random walks implies a strong notion of semantic similarity to construct embeddings -- examples include \textit{DeepWalk} \cite{perozzi2014deepwalk} and \textit{Node2Vec} \cite{grover2016node2vec}.

While the aforementioned approaches are effective in preserving network structure, semi-supervised learning with graph-structured data requires feature learning from node attributes in order to effectively propagate labels to unlabeled nodes. Since convolutional neural networks (CNNs) have been the mainstay for feature learning with data defined on regular-grids, a natural idea is to generalize convolutions to graphs. Existing work on this generalization can be categorized into \textit{spectral} approaches~\cite{bruna2013spectral,defferrard2016convolutional}, which operate on an explicit spectral representation of the graphs, and \textit{non-spectral} approaches that define convolutions directly on the graphs using spatial neighborhoods~\cite{duvenaud2015convolutional,niepert2016learning}. More recently, \textit{graph attention networks} (GAT) ~\cite{velickovic2017graph} have been introduced as an effective alternative for graph data modeling. An attention model parameterizes the local dependencies to determine the most relevant parts of the graph to focus on, while computing the features for a node. Unlike spectral approaches, attention models do not require an explicit definition of the Laplacian operator, and can support variable sized neighborhoods. However, a key assumption with these feature learning methods is that we have access to node attributes, in addition to the network structure, which is not the case in several applications.

\subsection{Proposed Work}
In this paper, we present a novel approach, \textit{GrAMME} (Graph Attention Models for Multi-layered Embeddings), for constructing multi-layered graph embeddings using attention models. In contrast to the existing literature on community detection, we propose to perform feature learning in an end-to-end fashion with the node classification objective, and show that it is superior to employing separate stages of network embedding (e.g. DeepWalk) and classifier design. First, we argue that even in datasets that do not have explicit node attributes, using random features is a highly effective choice. Second, we show that attention models provide a powerful framework for modeling inter-layer dependencies, and can easily scale to a large number of layers. To this end, we develop two architectures, \textit{GrAMME-SG} and \textit{GrAMME-Fusion}, that employ deep attention models for semi-supervised learning. While the former approach introduces virtual edges between the layers and constructs a \textit{Supra Graph} to parameterize dependencies, the latter approach builds layer-specific attention models and subsequently obtains consensus representations through fusion for label prediction. Using several benchmark multi-layered graph datasets, we demonstrate the effectiveness of random features and show that the proposed approaches significantly outperform state-of-the-art network embedding strategies such as DeepWalk. The main contributions of this work can be summarized as follows:
\begin{itemize}
	\item Existing multi-layered graph embedding approaches rely on parametric models to exploit dependencies between layers. In contrast, we propose a completely non-parametric solution based solely on attention models.
	\item We propose two attention-based architectures, GrAMME-SG and GrAMME-Fusion, that invoke information fusion at different stages of semi-supervised classification. 
	\item We introduce a weighted attention mechanism that is found to require much fewer number of attention heads to achieve the same performance as a regular GAT.
	\item We show that in cases where we do not have access to explicit node attributes, using random attributes is an effective choice.
	\item We evaluate our approaches on several benchmark datasets and demonstrate superior performance over state-of-the-art single-layered and multi-layered network embedding baselines. 
\end{itemize}

\section{Preliminaries}
\label{lab:prelims}
\vspace{-0.5mm}A single-layered undirected, unweighted graph is represented by $\mathcal{G} = (\mathcal{V}, \mathcal{E})$, where $\mathcal{V}$ denotes the set of nodes with cardinality $|\mathcal{V}| = N$, and $\mathcal{E}$ denotes the set of edges. A multi-layered graph is represented using a set of $L$ inter-dependent graphs $\mathcal{G}^{(l)} = (\mathcal{V}^{(l)}, \mathcal{E}^{(l)}), \text{for } l = 1,\dots,L$, where there exists a node mapping between every pair of layers to indicate which vertices in one graph correspond to vertices in the other. In our setup, we assume $\mathcal{V}^{(l)}$ from all layers contain the same set of nodes, while the edge sets $\mathcal{E}^{(l)}$ (each of cardinality $M^{(l)}$) are assumed to be different. In addition to the network structure, each node is endowed with a set of attributes, $\mathbf{x}_i\in \mathbb{R}^D$, $i \in [N]$, which can be used to construct latent representations, $\mathbf{Z} \in \mathbb{R}^{N \times d}$, where $d$ is the desired number of latent dimensions. Finally, each node is associated with a label $y_i$, which contains one of the $C$ pre-defined categories. Table \ref{tab:notation} summarizes all the notations used throughout this paper. In this paper, we consider the problem of performing label prediction at each of the nodes by exploiting the graph structure.

\begin{definition}{(\textit{Node classification})}
	Given a multi-layered network $\{(\mathcal{V}^{(l)}, \mathcal{E}^{(l)})\}_{l=1}^L$ and the semantic labels $\mathcal{Y}_{lab}$ for a subset of nodes $\mathcal{V}_{lab} \subset \mathcal{V}$, where each $y \in \mathcal{Y}_{lab}$ assumes one of the $C$ pre-defined classes, predict labels for each of the nodes in the set $v \in \mathcal{V} \setminus \mathcal{V}_{lab}$.
\end{definition}

\begin{table}[t]
	\centering
	\caption{Summary of the notations and their definitions.}
	\label{tab:notation}
	\begin{tabular}{p{1cm} p{6.5cm}}
		\hline
		\multicolumn{1}{c}{\textbf{Notation}} & \multicolumn{1}{c}{\textbf{Definition}}                                                                              \\ \hline \hline
		$\mathcal{V}$                           & Set of nodes in a graph                                                                                           \\
		$\mathcal{E}$                           & Set of edges in a graph                                                                                           \\
		$N$                                     & Number of nodes, $|\mathcal{V}|$                                       \\
		$L$                                     & Number of layers in a multi-layered graph                                                               \\
		$\mathcal{E}^{(l)}$                    & Edge set of $l^{th}$ layer of a multi-layered graph                          \\
		$M^{(l)}$								& Cardinality of the edge set $\mathcal{E}^{(l)}$ \\
		$\mathbf{x}_i$                          & Attributes for node $i$                                                                          \\
		$\mathbf{X}$                            &  Set of attributes for all $N$ nodes, $[\mathbf{x}_1, \mathbf{x}_2, \cdots , \mathbf{x}_N]^T$             \\
		$\mathbf{z}_i$                          & Embedding for the $i^{th}$ node                                                     \\
		$\mathbf{Z}$                            & Set of embeddings for all $N$ nodes, $[\mathbf{z}_1, \mathbf{z}_2, \cdots , \mathbf{z}_N]^T$ \\
		$D$                                     & Dimensionality of node attributes                                                                       \\
		$d$                                     & Embedding size                                                                       \\
		$y_i$                                   & Label for node $i$                                                                                        \\
		$\mathbf{W}$							& Learnable weight matrix for the linear transformation	\\
		$\mathbf{A}$                            & Parameters of the attention function                                                        \\
		$e_{ij}$                                & Attention coefficient for edge between $i$ \& $j$                                               \\
		$\alpha_{ij}$                           & Normalized attention coefficient for edge between $i$ \& $j$                                                           \\
		$H$                                     & Number of attention  heads                                                                                         \\
		$\beta_h$                               & Scaling factor for attention head $h$                                      \\
		$K$										& Number of supra-fusion heads									\\
		$\gamma^{(k)}$							& Scaling factor for the $k^{th}$ supra-fusion head									\\ \hline
%		$T$										& Number of graph attention layers    \\  \hline                    
	\end{tabular}
\end{table}

\subsection{Deep Network Embeddings}
\label{sec:netemb}
The scalability challenge of factorization techniques has motivated the use of deep learning methods to obtain node embeddings. The earliest work to report results on this direction was the DeepWalk algorithm by Perozzi \textit{et al.} \cite{perozzi2014deepwalk}. Interestingly, it draws analogy between node sequences generated by short random walks on graphs and sentences in a document corpus. Given this formulation, the authors utilize popular language modeling tools to obtain latent representations for the nodes \cite{mikolov2013distributed}. Let us consider a simple metric walk $\mathcal{W}_t$ in step $t$, which is rooted at the vertex $v_i$. The transition probability between the nodes $v_i$ and $v_j$ can be expressed as
\begin{equation}
P(\mathcal{W}_{t+1} = v_j | \mathcal{W}_t = v_i) = h(\|\mathbf{z}_i - \mathbf{z}_j\|_2 / \sigma),
\end{equation}where $\|\mathbf{z}_i - \mathbf{z}_j\|_2$ indicates the similarity metric between the two vertices in the latent space to be recovered and $h$ is a linking function that connects the vertex similarity to the actual co-occurrence probability. With appropriate choice of the walk length, the true metric can be recovered accurately from the co-occurrence statistics inferred using random walks. Furthermore, it is important to note that the frequency in which vertices appear in the short random walks follows a power-law distribution, similar to words in natural language. Given a length-$S$ sequence of words, $(w_0, w_1,\dots,w_{S-1})$, where $w_s$ denotes a word in the vocabulary, neural word embeddings attempt to obtain vector spaces that can recover the likelihood of observing a word given its context, i.e., $P(w_s | w_0, w_1,\dots,w_{s-1})$
over all sequences. Extending this idea to the case of graphs, a random walk on the nodes, starting from node $v_i$, produces the sequence analogous to sentences in language data. 

\subsection{Graph Attention Models}
\label{sec:gat}
In this section, we discuss the recently proposed graph attention model~\cite{velickovic2017graph}, a variant of which is utilized in this paper to construct multi-layered graph embeddings. Attention mechanism is a widely-adopted strategy in sequence-to-sequence modeling tasks, wherein a parameterized function is used to determine relevant parts of the input to focus on, in order to make decisions. A recent popular implementation of the attention mechanism in sequence models is the \textit{Transformer} architecture by Vaswani \textit{et al.}~\cite{vaswani2017attention}, which employs \textit{scalar dot-product} attention to identify dependencies. Furthermore, this architecture uses a \textit{self-attention} mechanism to capture dependencies within the same input and employs multiple \textit{attention heads} to enhance the modeling power. These important components have been subsequently utilized in a variety of NLP tasks~\cite{yang2017improving,barone2017deep} and clinical modeling~\cite{song2017attend}.

One useful interpretation of self-attention is that it implicitly induces a graph structure for a given sequence, where the nodes are time-steps and the edges indicate temporal dependencies. Instead of a single attention graph, we can actually consider multiple graphs corresponding to the different attention heads, each of which can  be  interpreted  to  encode  different  types  of  edges and hence can provide complementary information about different types of dependencies. This naturally motivates the use of attention mechanism in modeling graph-structured data. Recently, Velickovic \textit{et al.} generalized the idea in~\cite{velickovic2017graph} to create a graph attention layer, that can be stacked to build effective deep architectures for semi-supervised learning tasks. In addition to supporting variabilities in neighborhood sizes and improving the model capacity, graph attention models are computationally more efficient than other graph convolutional networks. In this paper, we propose to utilize attention mechanisms to model multi-layered graphs.

\noindent \textbf{Formulation}: A head in the graph attention layer learns a latent representation for each node by aggregating the features from its neighbors. More specifically, the feature at a node is computed as the weighted combination of features from its neighbors, where the weights are obtained using the attention function. Following our notations, each node $v_i$ is endowed with a $D-$dimensional attribute vector $\mathbf{x}_i$, and hence the input to graph attention layer is denoted by the set of attributes $\{\mathbf{x}_1,\mathbf{x}_2, \cdots, \mathbf{x}_n\}$. The attention layer subsequently produces $d-$dimensional latent representations $\mathbf{Z} = \{\mathbf{z}_1,\mathbf{z}_2, \cdots, \mathbf{z}_n\}$. 

An attention head is constructed as follows: First, a linear transformation is applied to the features at each node, using a shared and trainable weight matrix $\mathbf{W} \in \mathbb{R}^{d\times D}$, thus producing intermediate representations,
\begin{equation}
\label{eq:one}
\mathbf{\tilde{X}} = \mathbf{X}  \mathbf{W^T}.
\end{equation}Subsequently, a scalar dot-product attention function is utilized to determine attention weights for every edge in the graph, based on features from the incident neighbors. Formally, the attention weight for the edge $e_{ij}$ connecting the nodes $v_i$ and $v_j$ is computed as
\begin{equation}
\label{eq:two}
e_{ij} = \langle \mathbf{A}, \tilde{\mathbf{x}}_i || \tilde{\mathbf{x}}_j \rangle,
\end{equation}where $\mathbf{A}\in \mathbb{R}^{2d \times 1}$ denotes the parameters of the attention function, and $||$ represents concatenation of features from nodes $v_i$ and $v_j$ respectively. The attention weights $e_{ij}$ are computed with respect to every node in the neighborhood of $v_i$, i.e., for $v_j \in \mathcal{N}_i\cup\{i\}$, where $\mathcal{N}_i$ represents the neighborhood of $v_i$. Note that, we include the self-edge for every node while implementing the attention function. The weights are then normalized across all neighboring nodes using a softmax function, thus producing the normalized attention coefficients.
\begin{equation}
\alpha_{ij} = \text{softmax}(e_{ij})
\end{equation}Finally, the normalized attention coefficients are used to compute the latent representation at each node, through a weighted combination of the node features. Note that, a non-linearity function is also utilized at the end to improve the approximation. The optimal values for model parameters $\mathbf{W}$ and $\mathbf{A}$ are obtained through backpropagation in an end-to-end learning.
\begin{equation}
\label{eqn:attapp}
\mathbf{z}_i = \sigma \Bigg( \sum_{j \in \mathcal{N}_i\cup\{i\} } \alpha_{ij}  \tilde{\mathbf{x}_j}\Bigg)
\end{equation}  An important observation is that the attention weights are not required to be symmetric. For example, if a node $v_i$ has a strong influence on node $v_j$, it does not imply that node $v_j$ also has a strong influence on $v_i$ and hence $e_{ij} \not= e_{ji}$. The operations from equations (\ref{eq:one}) to (\ref{eqn:attapp}) constitute a single head. While this simple parameterization enables effective modeling of relationships in a graph while learning latent features, the modeling capacity can be significantly improved by considering multiple attention heads. Following the \textit{Transformer} architecture~\cite{vaswani2017attention}, the output latent representations from the different heads can be aggregated using either concatenation or averaging operations.

\section{Proposed Approaches}
\label{sec:approach}
\begin{figure}[t]
	\centering
	\includegraphics[width=0.9\linewidth]{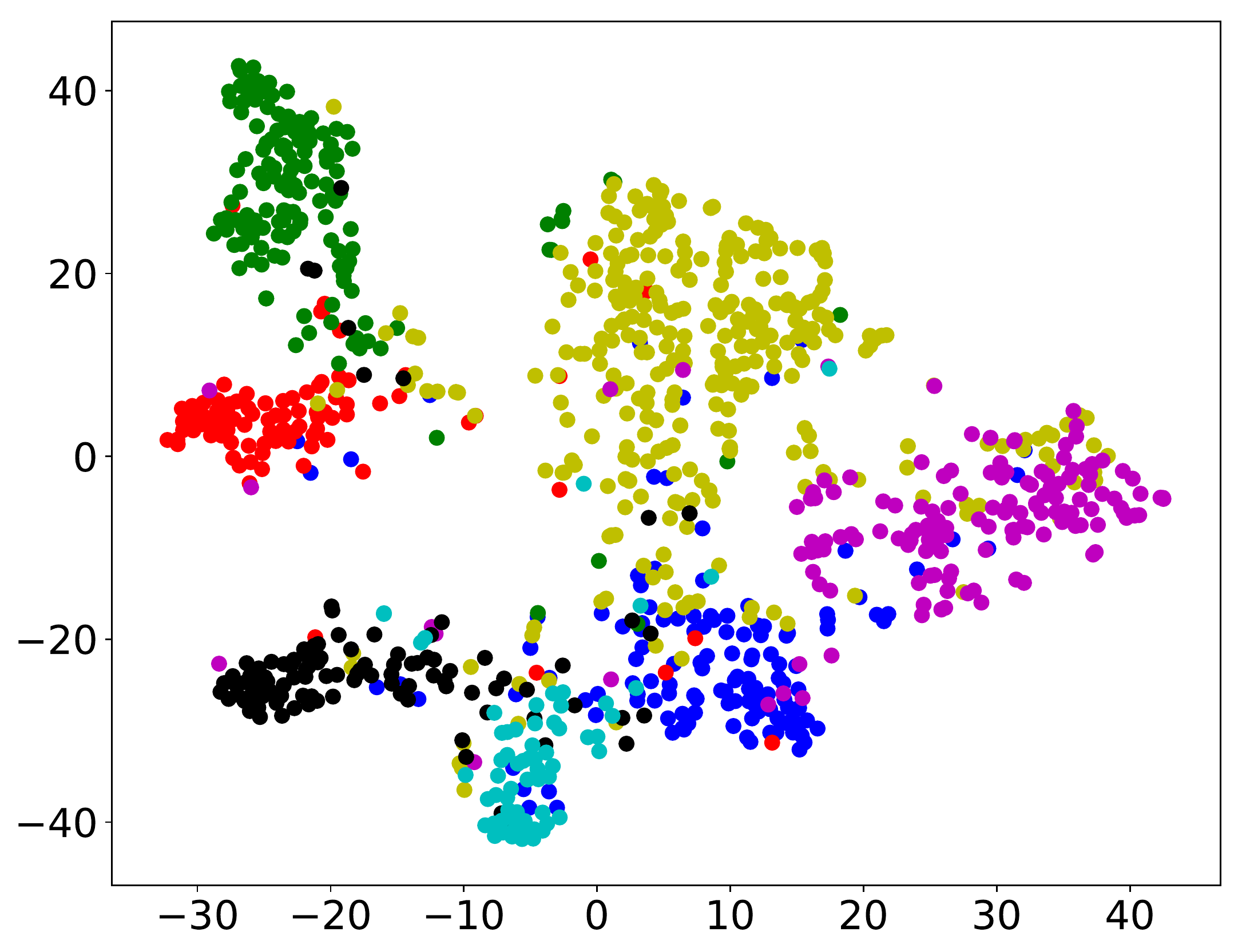}
	\caption{$2-$D Visualization of the embeddings for the single-layer \textit{Cora} dataset obtained using the proposed weighted attention mechanism.}
	\label{fig:cora_embeddings}
\end{figure}

\begin{figure*}[t]
	\centering
	\includegraphics[width=\linewidth]{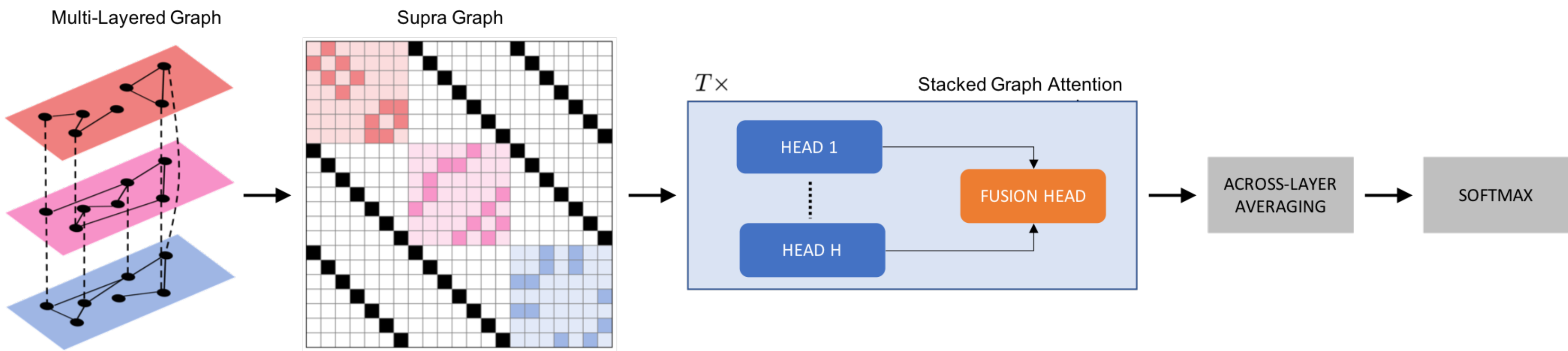}
	\caption{\textit{GrAMME-SG} Architecture: Proposed approach for obtaining multi-layered graph embeddings with attention models applied to the \textit{Supra Graph}, constructed by introducing virtual edges between layers.}
	\label{fig:gramme-sg}
\end{figure*}

In this section, we discuss the two proposed approaches for constructing multi-layered graph embeddings in semi-supervised learning problems. Before presenting the algorithmic details, we describe the attention mechanism used in our approach, which utilizes a weighting function to deal with multiple attention heads. Next, we motivate the use of randomized node attributes for effective feature learning. As described in Section \ref{sec:intro}, in multi-layered graphs, the relationships between nodes are encoded using multiple edge sets. Consequently, while applying attention models for multi-layered graphs, a node $v_i$ in layer $l$ needs to update its hidden state using not only knowledge from its neighborhood in that layer, but also the shared information from other layers. Note, we assume no prior knowledge on the dependency structure, and solely rely on attention mechanisms to uncover the structure.

\subsection{Weighted Attention Mechanism}
\label{ssec:fusion_head}
From the discussion in Section \ref{sec:gat}, it is clear that latent representations from the multiple attention heads can provide complementary information about the node relationships. Hence, it is crucial to utilize that information to produce reliable embeddings for label propagation. When simple concatenation is used, as done in~\cite{velickovic2017graph}, an attention layer results in features of dimension $H \times d$, where $H$ is the number of attention heads. While this has been effective, one can gain improvements by performing a weighted combination of the attention heads, such that different heads can be assigned varying levels of importance. This is conceptually similar to the \textit{Weighted Transformer} architecture proposed by Ahmed \textit{et al.}~\cite{2017Karim}. For a node $v_i$, denoting the representations from the different heads as $\mathbf{z}_i^1 \cdots \mathbf{z}_i^H$, the proposed weighted attention combines these representations as follows:
\begin{equation}
\hat{\mathbf{z}}_i = \sum_{h=1}^H \beta_h \mathbf{z}_i^h,
\label{eqn:watt}
\end{equation}where $\beta_h$ denotes the scaling factor for head $h$ and are trainable during the optimization. Note that, the scaling factors are shared across all nodes and they are constrained to be non-negative. Optionally, one can introduce the constraint $\sum_h \beta_h = 1$ into the formulation. However, we observed that its inclusion did not result in significant performance improvements in our experiments. Given a set of attention heads for a single graph layer, we refer to this weighting mechanism as a \textit{fusion head}.

Interestingly, we find that this modified attention mechanism produces robust embeddings, when compared to the graph attention layer proposed in~\cite{velickovic2017graph}, even with lesser number of attention heads. For example, let us consider Cora, a single-layered graph dataset, containing $2708$ nodes (publications) belonging to one of $7$ classes. With the regular graph attention model, comprised of two attention layers with $8$ heads each, we obtained a test accuracy of $81.5\%$ ($140$ training nodes). In contrast, our weighted attention, even with just $2$ heads, produces state-of-the-art accuracy of $82.7\%$. Naturally, this leads to significant reduction in the computational complexity of our architecture, which is more beneficial when dealing with multi-layered graphs. Figure \ref{fig:cora_embeddings} illustrates a $2-$D visualization (obtained using t-SNE) of the embeddings from our graph attention model. 

\subsection{Using Randomized Node Attributes}
With graph attention models, it is required to have access to node attributes (or features), which are then used to obtain the latent representations. However, in many cases, multi-layered graph datasets are often comprised of only the edge sets, without any additional information. This scenario arises in several application domains including social networks, healthcare, transporation etc. For example in brain networks, nodes correspond to functional regions in the brain, while edges indicate correlations between activations in those regions. Similarly in air-transportation networks, each node corresponds to a city while an edge indicates whether there is a direct flight connection between cities. Consequently, in existing graph inferencing approaches (e.g. community detection), it is typical to adopt an unsupervised network embedding strategy, where the objective is to ensure that the learned representations preserve the network topology (i.e. neighborhoods). However, such an approach is not optimal for semi-supervised learning tasks, since the model parameters can be more effectively tuned using the task-specific objective, in an end-to-end fashion. In order to address this challenge, we propose to employ a randomized initialization strategy for creating node attributes. 

Interestingly, random initialization has been highly successful in creating word representations for NLP tasks, and in many scenarios its performance matches or even surpasses pre-trained word embeddings. With this initialization, the graph attention model can be used to obtain latent representations that maximally support label propagation in the input graph. Unlike fully supervised learning approaches, the embeddings for nodes that belong to the same class can still be vastly different, since the attention model fine-tunes the initial embeddings using only the locally connected neighbors. As we will show in our experiments, this simple initialization is effective, and our end-to-end training approach produces superior performance.

\subsection{Approach Description: GrAMME-SG}\label{ssec:sup-approach}
In this approach, we begin with the initial assumption that information is shared between all layers in a multi-layered graph, and use attention models to infer the actual dependencies, with the objective of improving label propagation performance. More specifically, we introduce virtual edges (also referred as pillar edges~\cite{kim2015community}) between every node in a layer and its counterparts in other layers, resulting in a supra graph, $\mathcal{G}_{sup}$. The block diagonals of the adjacency matrix for $\mathcal{G}_{sup}$ contains the individual layers, while the off-diagonal entries indicate the inter-layer connectivities. As illustrated in Figure \ref{fig:gramme-sg}, the virtual edges are introduced between nodes with the same ID across layers. This is a popularly adopted strategy in the recent community detection approaches~\cite{song2018}, however, with a difference that the nodes across layers are connected only when they share similar neighborhoods. In contrast, we consider all possible connections for information flow, and rely on the attention model to guide the learning process. Note that, it is possible that some of the layers can only contain a subset of the nodes. 

%construct a large supra graph with all available $P$ edge sets since each edge set encompasses specific relationship and we want to use all the information from the multi-layer graph data. We construct a supra-graph in which block diagonals of the supra graph's adjacency matrix represent the original adjacency matrix for the multi-layer graph. We also introduce edges between a node and its corresponding counterparts in different layers; construction of a supra-graph in the aforementioned manner is typically seen in community detection for a multi-layer graph. We construct the supra graph as follows. For node $i$ in layer $1$, we first add edges to its counterpart nodes from all the layers ($2$ to $P$). That is, we add edges, from node $i^{(1)}$ to $i^{(2)}$, $i^{(1)}$ to $i^{(3)}$, $\cdots$, $i^{(1)}$ to $i^{(P)}$. The superscript $(\cdot)$ denotes the graph-layers. In literature, these edge connections are often referred as "pillar" edges. Originally the multi-layer graph can have a maximum of $n^2P$ edges. Addition of pillar edges increases the maximum number of edges to $n^2P + nP^2$. Also, originally the graph consisted of $n$ nodes only. 

Following this, we generate random features of dimension $D$ at each of the nodes in $\mathcal{G}_{sup}$ and build a stacked attention model for feature learning and label prediction. Our architecture is comprised of $T$ graph attention layers, which in turn contains $H$ attention heads and a fusion head to combine the complementary representations. As discussed earlier, an attention head first performs a linear transformation on the input features, and parameterizes the neighborhood dependencies to learn locally consistent features. The neighborhood size for each node can be different, and we also include self edges while computing the attention weights. Since we are using the supra graph in this case, the attention model also considers nodes from the other layers. This exploits the inter-layer dependencies and produces latent representations that can be influenced by neighbors in the other layers. Following the expression in Equation (\ref{eqn:attapp}), the latent feature at a node $v_i$ in layer $l$ can be obtained using an attention head as follows:
\begin{equation}
\label{eqn:attapp_m}
\mathbf{z}_{i^{(l)}} = \sigma \Bigg( \sum_{j \in \mathcal{N}_{i^{(l)}}\cup\{i^{(1)} \cdots i^{(L)}\} } \alpha_{{i^{(l)}}j}  \tilde{\mathbf{x}_j}\Bigg),
\end{equation}where $\tilde{\mathbf{x}}_j$ denotes the linear-transformed feature vector for a node. This is repeated with $H$ attention heads with different parameters, and subsequently a fusion head is used to combine those representations. Note that, a fusion head is defined using $H$ scaling factors, denoting the importance for each of the heads. This operation can be formally stated as follows:
\begin{equation}
\hat{\mathbf{z}}_{i^{(l)}} = \sum_{h=1}^H \beta_h \mathbf{z}_{i^{(l)}}^h.
\label{eqn:watt_m}
\end{equation}Consequently, we obtain latent features of dimension $d$ for each node in $\mathcal{G}_{sup}$, which are then sequentially processed using additional graph attention layers. Since the overall goal is to obtain a single label prediction for each node, there is a need to aggregate features for a node from different layers. For this purpose, we perform an across-layer average pooling and employ a feed-forward layer with softmax activation for the final prediction.

\begin{figure}[t]
	\centering
	\includegraphics[width=\linewidth]{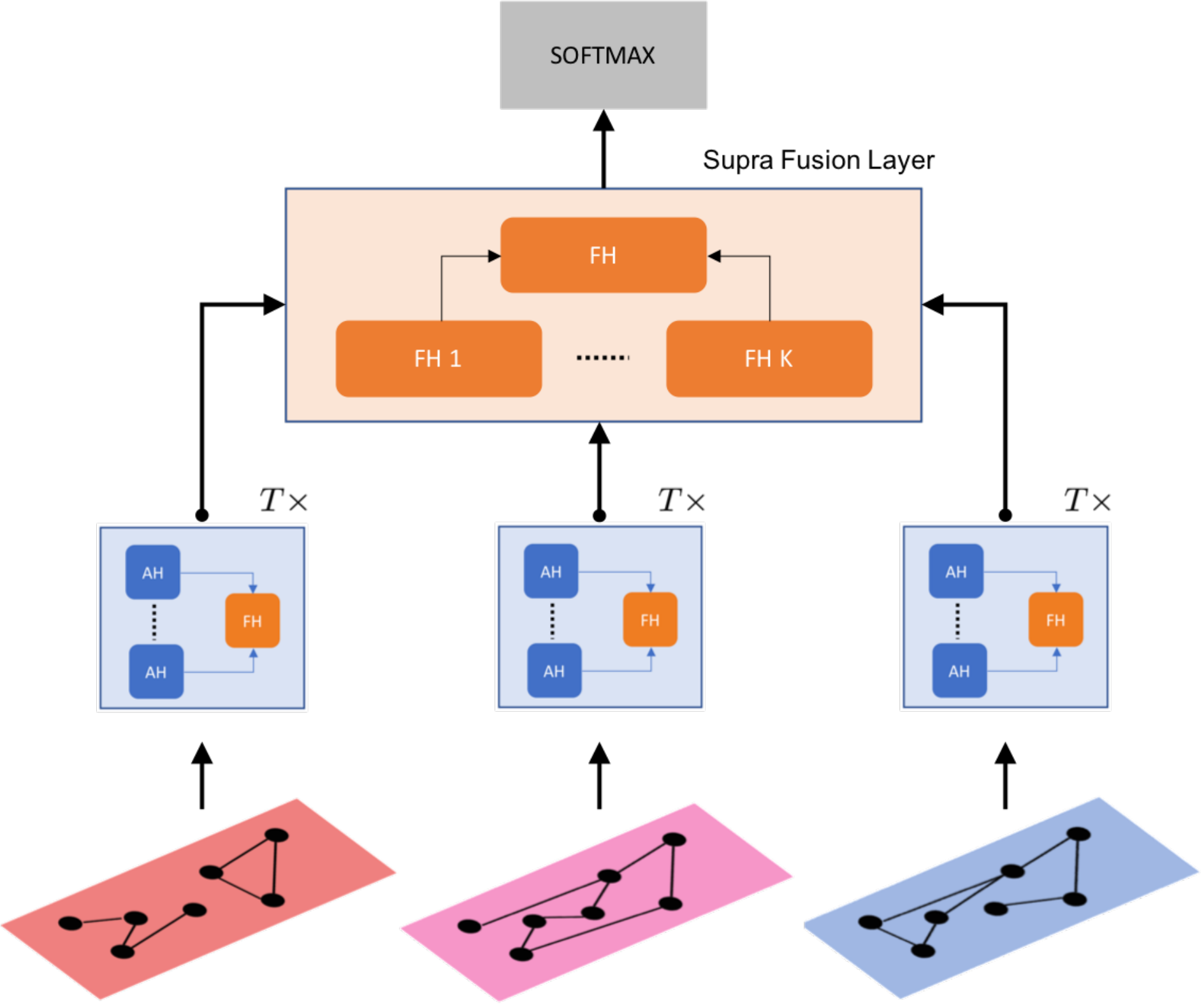}
	\caption{\textit{GrAMME-Fusion} Architecture: Proposed approach for obtaining multi-layered graph embeddings through fusion of representations from layer-wise attention models.}
	\label{fig:gramme-fusion}
\end{figure}

\begin{figure*}[t]
	\centering
	\subfigure[$10\%$ Train Nodes]{\includegraphics[width=.28\linewidth]{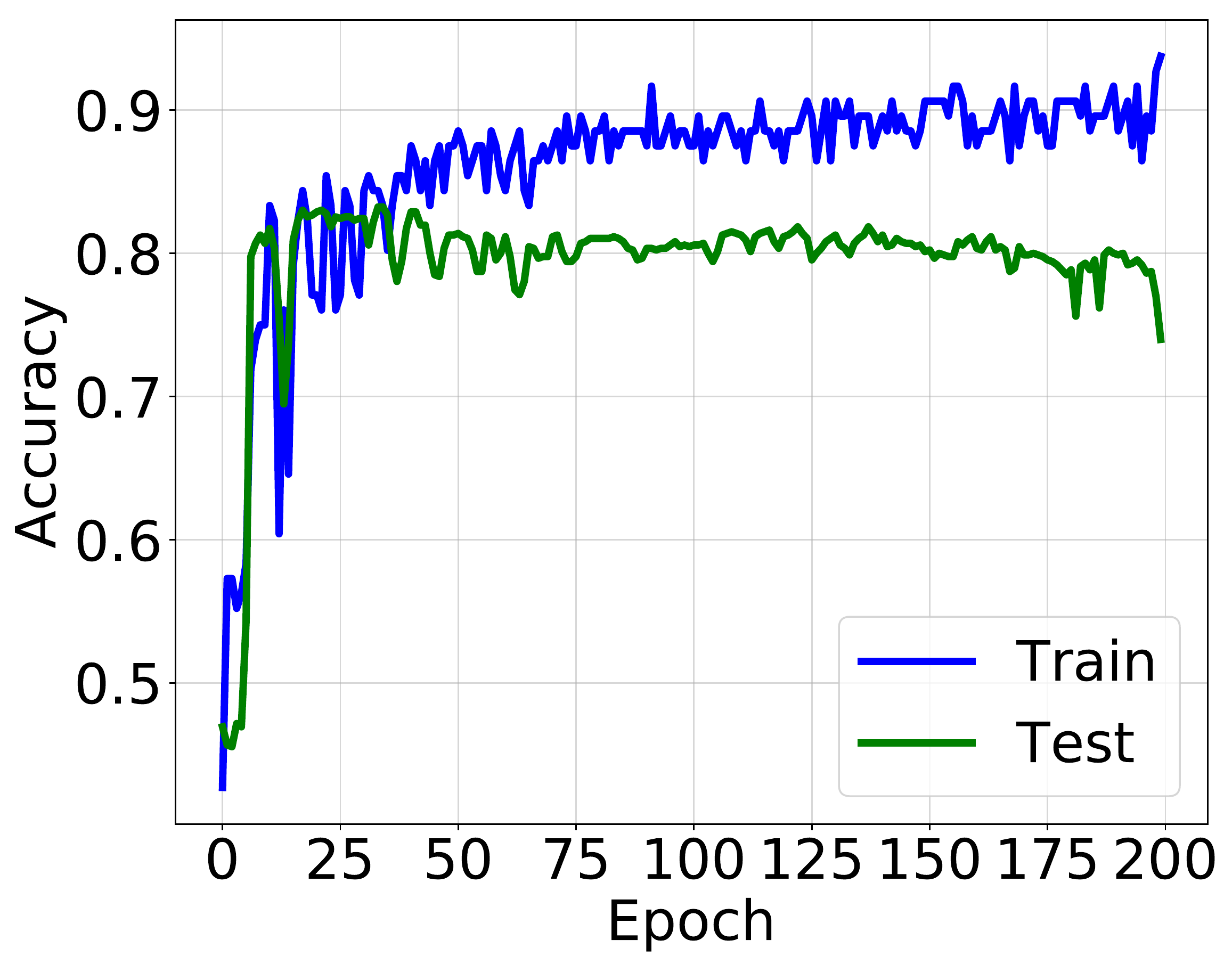}}
	\subfigure[$20\%$ Train Nodes]{\includegraphics[width=.28\linewidth]{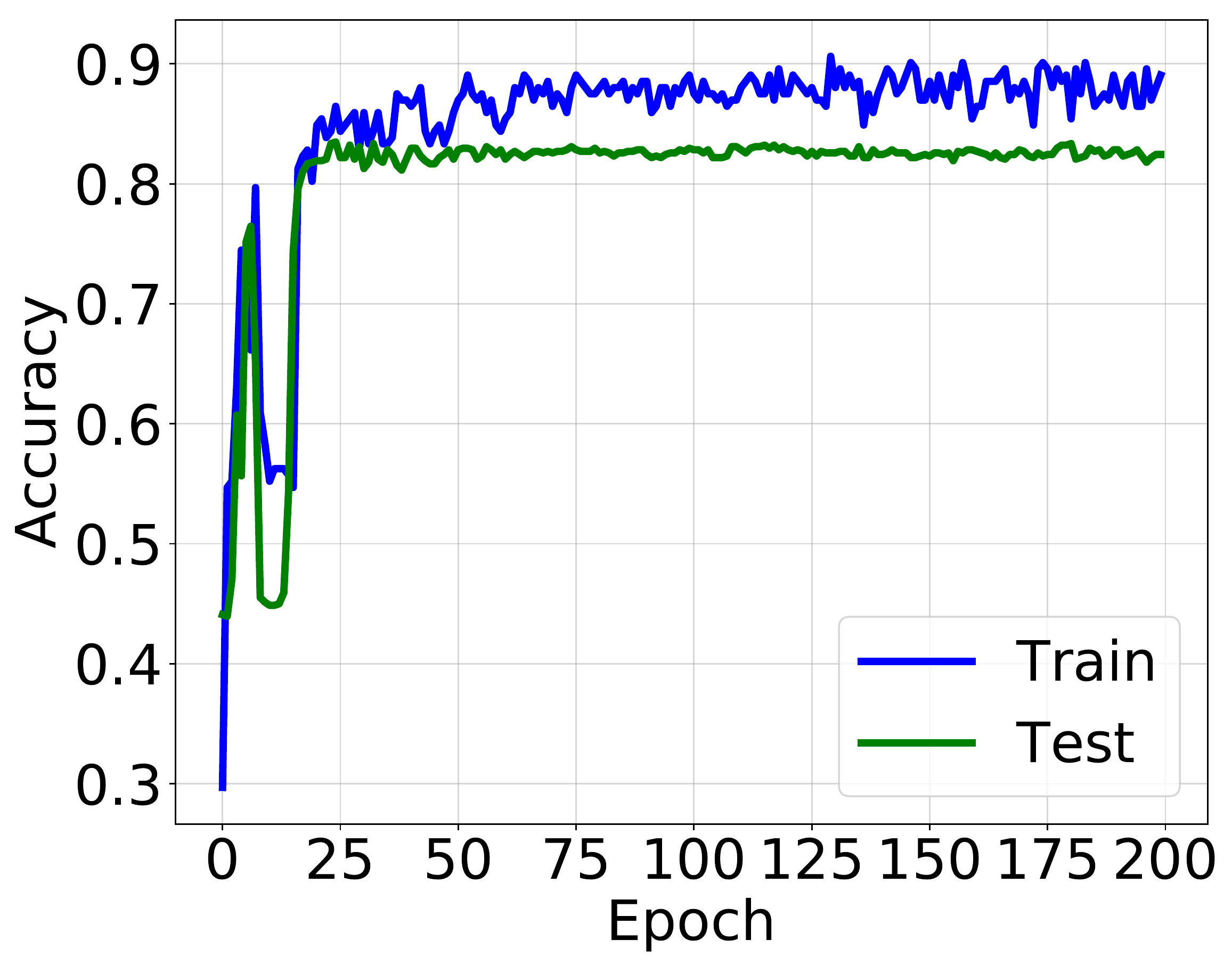}}
	\subfigure[$30\%$ Train Nodes]{\includegraphics[width=.28\linewidth]{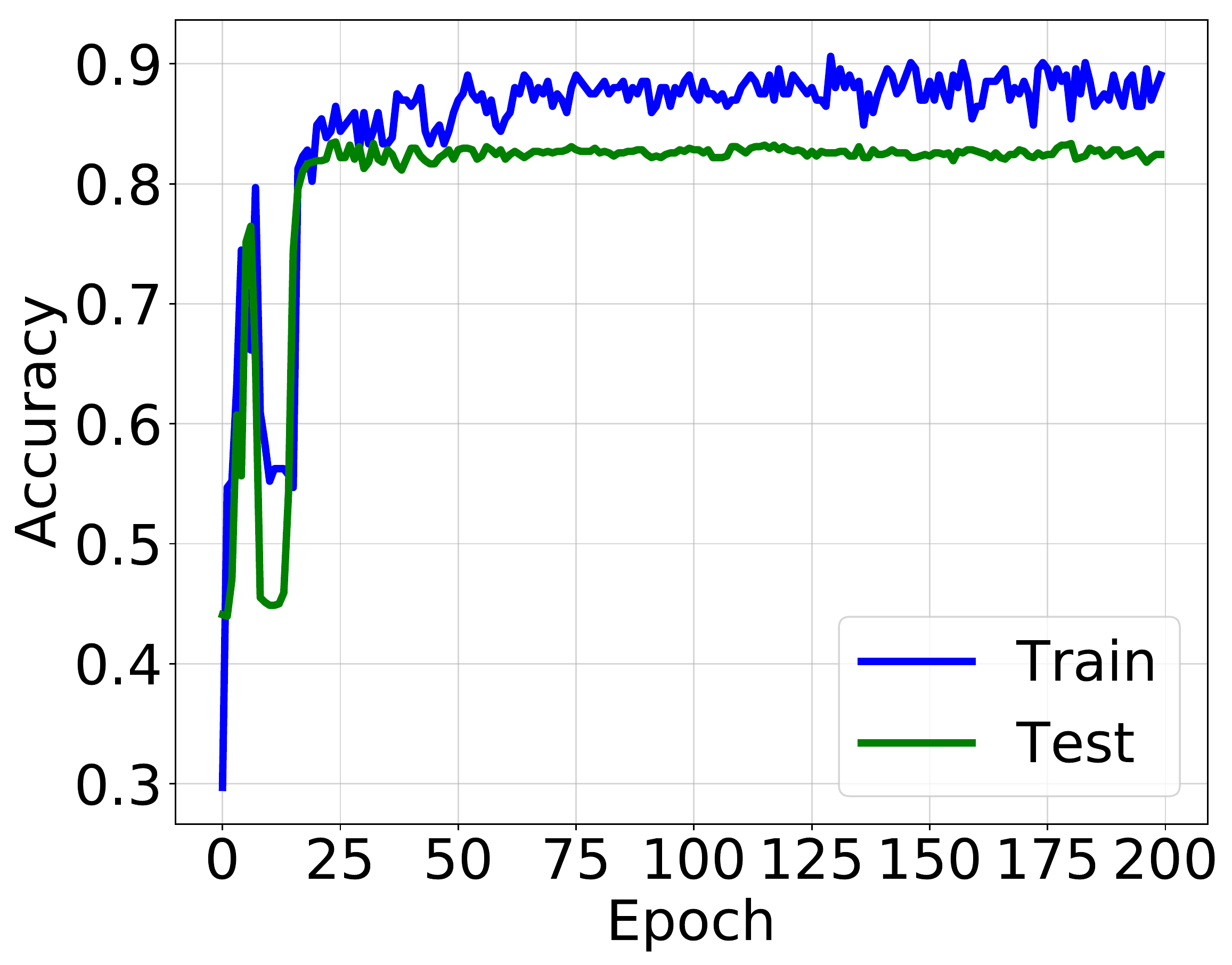}}
	
	\caption{Convergence characteristics of the proposed \textit{GrAMME-Fusion} architecture with the parameters $T = 2$, $H = 1$ and $K = 5$ respectively.}
	\label{fig:training}
\end{figure*} 

\subsection{Approach Description: GrAMME-Fusion}\label{ssec:sup-fusion}
While the \textit{GrAMME-SG} approach provides complete flexibility in dealing with dependencies, the complexity of handling large supra graphs is an inherent challenge. Hence, we introduce another architecture, \textit{GrAMME-Fusion}, which builds only layer-wise attention models, and introduces a \textit{supra fusion} layer that exploits inter-layer dependencies using only fusion heads. As described in Section \ref{ssec:fusion_head}, a fusion head computes simple weighted combination and hence is computationally cheap. For simplicity, we assume that the same attention model architecture is used for every layer, although that is not required. This approach is motivated from the observation that attention heads in our feature learning architecture, and the different layers in a multi-layered graph both provide complementary views of the same data, and hence they can be handled similarly using fusion heads. In contrast, \textit{GrAMME-SG} considers each node in every layer as a separate entity. Figure \ref{fig:gramme-fusion} illustrates the \textit{GrAMME-Fusion} architecture.

Initially, each graph layer $l$ is processed using an attention model comprised of $T$ stacked graph attention layers, each of which implements $H$ attention heads and a fusion head, to construct layer-specific latent representations. Though the processing of the $L$ layers can be parallelized, the computational complexity is dominated by the number of heads $H$ in each model. Next, we construct a \textit{supra fusion} layer which is designed extensively using fusion heads in order to parameterize the dependencies between layers. In other words, we create $K$ fusion heads with scaling factors $\boldsymbol{\gamma}^{(k)} \in \mathbb{R}^L, \forall k = 1 \cdots K$, in order to combine the representations from the $L$ layer-specific attention models. Note that, we use multiple fusion heads to allow different parameterizations for assigning importance to each of the layers. This is conceptually similar to using multiple attention heads. Finally, we use an overall fusion head, with scaling factors $\boldsymbol{\kappa} \in \mathbb{R}^K$, to obtain a consensus representation from the multiple fusion heads. One can optionally introduce an additional feed-forward layer prior to employing the overall fusion to improve the model capacity. The output from the \textit{supra fusion} layer is used to make the prediction through a fully-connected layer with softmax activation. The interplay between the parameters $H$ (layer-wise attention heads) and $K$ (fusion heads in the \textit{supra fusion} layer) controls the effectiveness and complexity of this approach.

\section{Empirical Studies}
\label{sec:results}
In this section, we evaluate the proposed approaches by performing semi-supervised node classification with several benchmark multi-layered graph datasets from diverse application domains. Our experiments study the behavior of our approaches and baseline methods in semi-supervised learning, with varying amounts of labeled nodes. Though the proposed approaches can be utilized for inductive learning, we restrict our experiments to transductive settings. Before presenting the detailed evaluation, we first describe the datasets considered for our study, and briefly discuss the baseline techniques.

\subsection{Datasets}
We describe in detail the multi-layered graph datasets used for evaluation. A summary of the datasets can be found in Table \ref{tab:dataset}.

\noindent \textbf{(i) Vickers-Chan}: The Vickers-Chan \cite{vickers1981representing} dataset represents the social structure of students from a school in Victoria, Australia. Each node represents a student studying in 7th grade, and the three graph layers are constructed based on student responses for three questions. The dataset is comprised of $29$ nodes and their gender value is used as the label in our learning formulation. 

\noindent \textbf{(ii) Congress Votes}: The Congress votes \cite{schlimmer1987concept} dataset is obtained from the 1984 United States Congressional Voting Records Database. This includes votes from every congressmen on $16$ different bills, which results in a $16$-layered graph. There are $435$ congressmen who are labeled as either democrats or republicans. For every layer, we establish an edge between two nodes in the corresponding layer, if those two congressmen voted similarly (``yes" or ``no"). 

\noindent \textbf{(ii) Leskovec-Ng}: This dataset \cite{chen2017multilayer} is a temporal collaboration network of professors Jure Leskovec and Andrew Ng. The $20$ year co-authorship information is partitioned into $5$-year intervals, in order to build a $4$-layered graph. In any layer, two researchers are connected by an edge if they co-authored at least one paper in the $5$-year interval. Each researcher is labeled as affiliated to either Leskovec's or Ng's group.

\noindent \textbf{(iv) Reinnovation}: This dataset describes the Global Innovation Index for $144$ countries (nodes). For each node, the label represents the development level of that corresponding country. There are $3$ levels of development, thus representing the $3$ classes. Each layer in a graph is constructed based on similarities between countries in different sectors. The network contains $12$-layers from obtained from $12$ sectors that include infrastructure, institutions, labor market, financial market etc.

\noindent \textbf{(v) Mammography (UCI)}: This dataset contains information about mammographic mass lesions from $961$ subjects. We consider different attributes, namely the BI-RADS assessment, subject age, shape, margin, and density of the lesion, in order to construct the different layers of the graph. This data is quite challenging due to the presence of $~2$ million edges. Conventional network embedding techniques that rely on sparsity of the graphs can be particularly ineffective in these scenarios. Finally, the lesions are either marked as benign or malignant, to define the labels.

\noindent \textbf{(vi) CKM}: This dataset \cite{coleman1957diffusion} is collected from physicians in four different towns (classes). This dataset models information diffusion through social networks among the physicians. There are three layers which encode similarity in the responses to a questionnaire.

\noindent \textbf{(vii) Balance Scale}: This dataset \cite{Dua:2019} summarizes the results from a psychological experiment. Using $4$ different attributes characterizing a subject namely left weight, the left distance, the right weight, and the right distance, we constructed a $4-$layered graph. Each subject is classified as having the balance scale tip to the right, tip to the left, or be balanced.

\subsection{Baselines}
Given that the datasets considered do not contain specific node attributes to perform feature learning, the natural approach is to obtain embeddings for each node, and to subsequently build a classifier model. We compare our proposed architectures with the following state-of-the-art single-layered and multi-layered graph embedding techniques.
	
\noindent \textbf{DeepWalk}: DeepWalk \cite{perozzi2014deepwalk} is a random-walk based embedding technique that uses a deep neural network. Random walks on a graph is analogous to sentences in a document, and hence co-occurring nodes are embedded together.

\noindent \textbf{Node2Vec}: Node2Vec \cite{grover2016node2vec} is similar to DeepWalk, but it introduces bias in random walks with two additional parameters that trade-off between depth-first and breadth-first walks.

\noindent \textbf{LINE}: LINE \cite{tang2015line} is similar to DeepWalk but adds information from second hop friends in its random walks. This enables nodes with shared neighborhoods to have similar embeddings.

\noindent \textbf{PMNE}: This method~\cite{liu2017principled} uses different merge strategies to combine embeddings from each of the layers in a multi-layered network. We consider the results aggregation strategy, since it often outperforms other variants

\noindent \textbf{MNE}: This recent multiplex network embedding \cite{zhang2018scalable} technique uses a unified network embedding model that generates, for each node, a high-dimensional common embedding and low dimensional embedding for each aspect of relationship.

%We use the two following baselines in order to compare the performance of the proposed approaches. Given that the datasets considered do not contain specific node attributes to perform feature learning, the natural approach is to obtain embeddings for each node in every layer, using deep network embedding techniques, and to subsequently build a classifier model using the resulting features. Following recent approaches such as \cite{li2018multi}, we choose DeepWalk, which is a state-of-the-art embedding technique, for obtaining deep embeddings. In particular, we consider two different variants: (i) \textit{DeepWalk}: We treat each layer in the multi-layered graph as independent, and obtain embeddings from the layers separately. Finally, we concatenate the embeddings for each node from the different layers and build a multi-layer perceptron to perform the classification; (ii) \textit{DeepWalk-SG}: We construct a supra graph by introducing virtual edges between nodes across layers (as described in Section \ref{ssec:sup-approach}) and perform DeepWalk on the supra graph. Finally, the embeddings are concatenated as in the previous case and the classifier is designed. Though the former approach does not exploit the inter-layer information, in cases where there is significant variability in neighborhood structure across layers, it can still be effective by treating the layers independently.
\begin{table*}[t]
	\centering
	\caption{Summary of the datasets used in our empirical studies.}
	\label{tab:dataset}
	\begin{tabular}{|c|c|c|c|c|c|}
		\hline
		\textbf{Dataset}        & \textbf{Type}            & \textbf{\# Nodes} & \textbf{\# Layers} & \textbf{\# Total edges} & \textbf{\# Classes} \\ \hline \hline
		Vickers-Chan       & Classroom social structure & 29    & 3  &740              & 2       \\
		Congress Votes & Bill voting structure among senators        & 435   & 16      & 358,338          & 2       \\
		Leskovec-Ng    & Academic collaboration     & 191   & 4      & 1,836          & 2       \\
		Reinnovation   & Global innovation index similarities       & 145   & 12   & 18,648           & 3 \\
		Mammography	   & Mammographic Masses	& 961 & 5 & 1,979,115  & 2   \\
		CKM(Social)	   & Responses to Questions  				& 241   & 3 & 3825 		& 4\\	
		Balance Scale  & Psychological assessment & 625    & 4    & 312,500   & 3 	   \\ \hline
	\end{tabular}
\end{table*}

\begin{figure*}[th]
	\centering
	\subfigure[Congress Votes -- Initial]{\includegraphics[width=.24\linewidth]{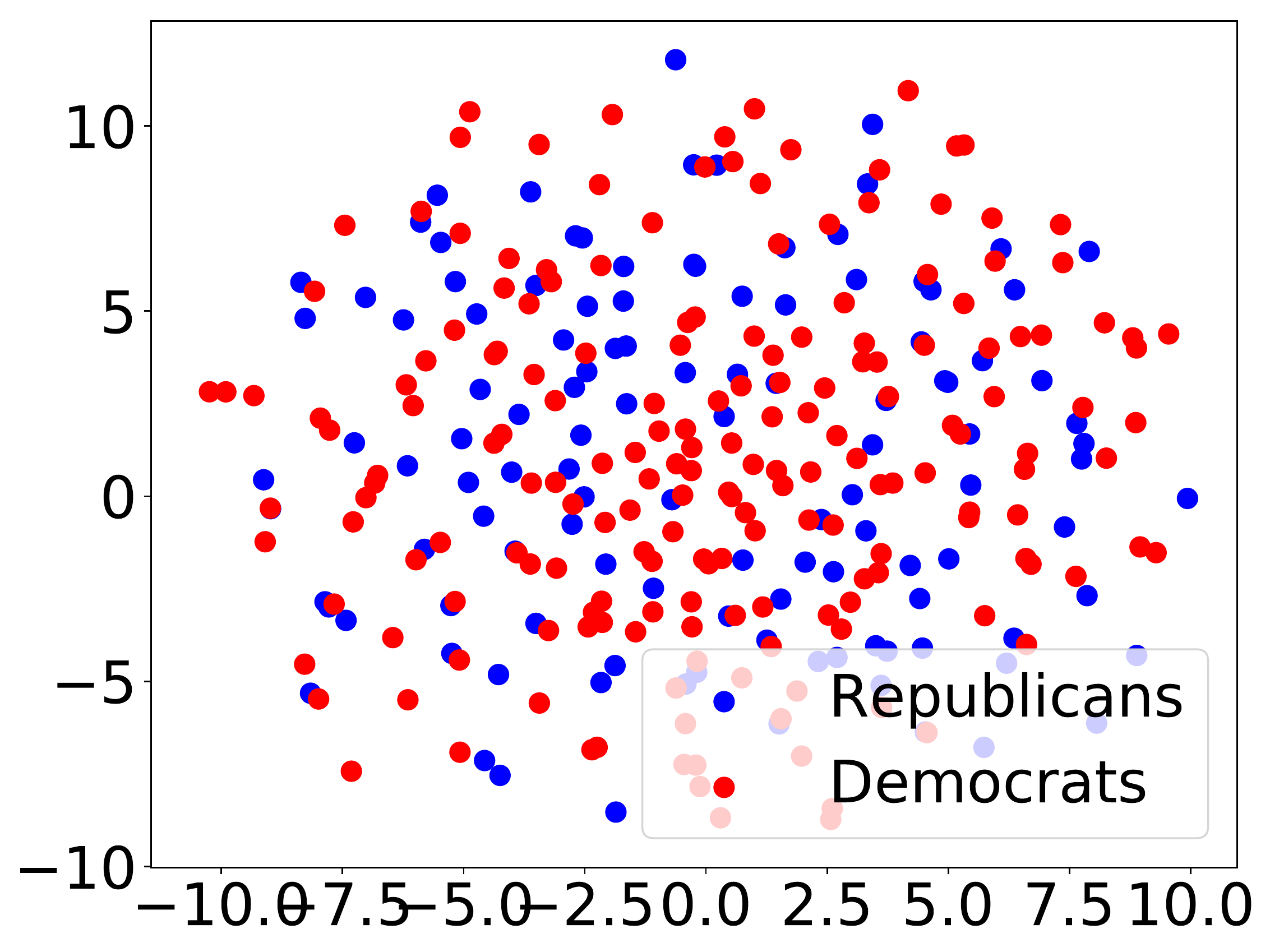}}
	\subfigure[Congress Votes -- Final]{\includegraphics[width=.24\linewidth]{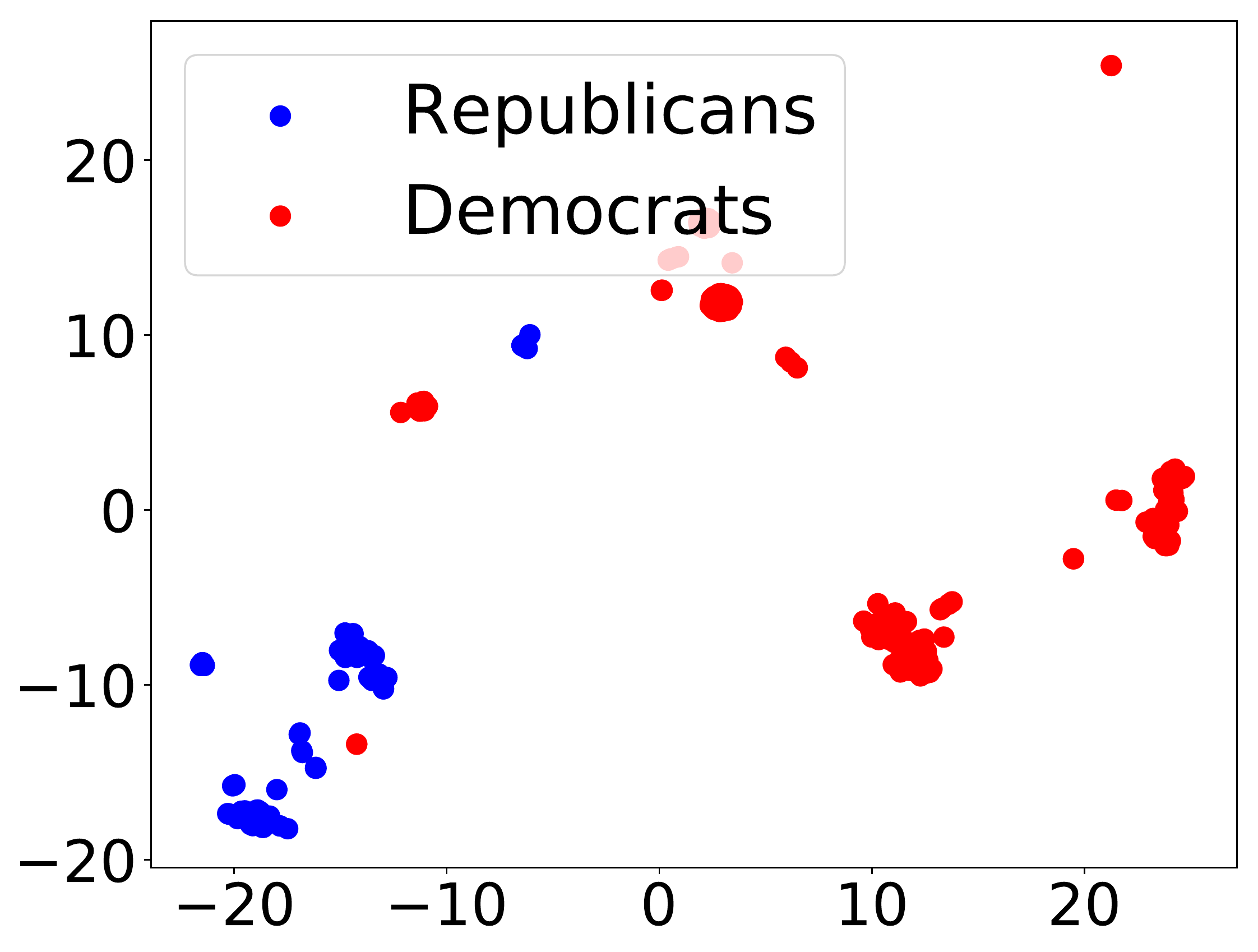}}
	\subfigure[Mammography -- Initial]{\includegraphics[width=.24\linewidth]{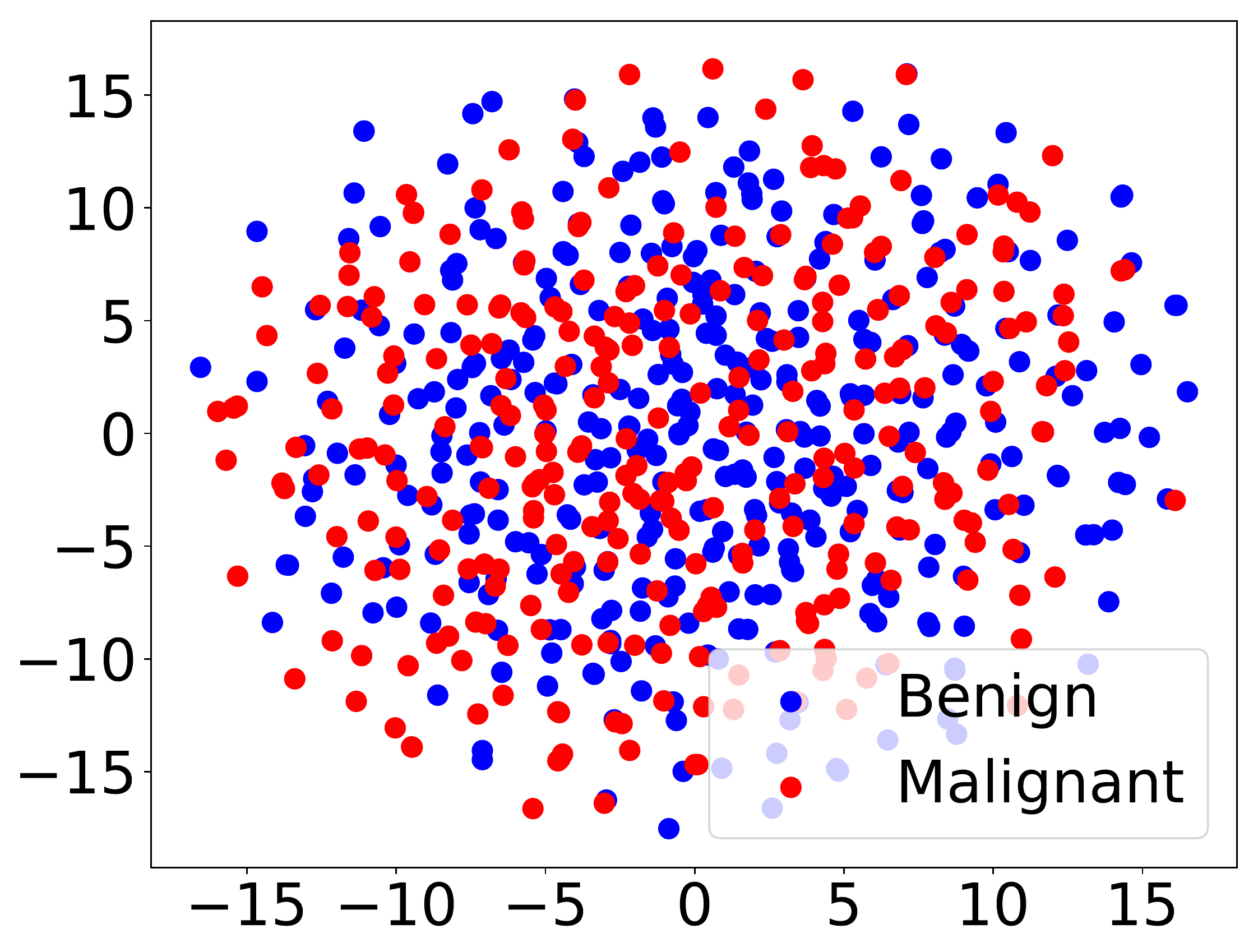}}
	\subfigure[Mammography -- Final]{\includegraphics[width=.24\linewidth]{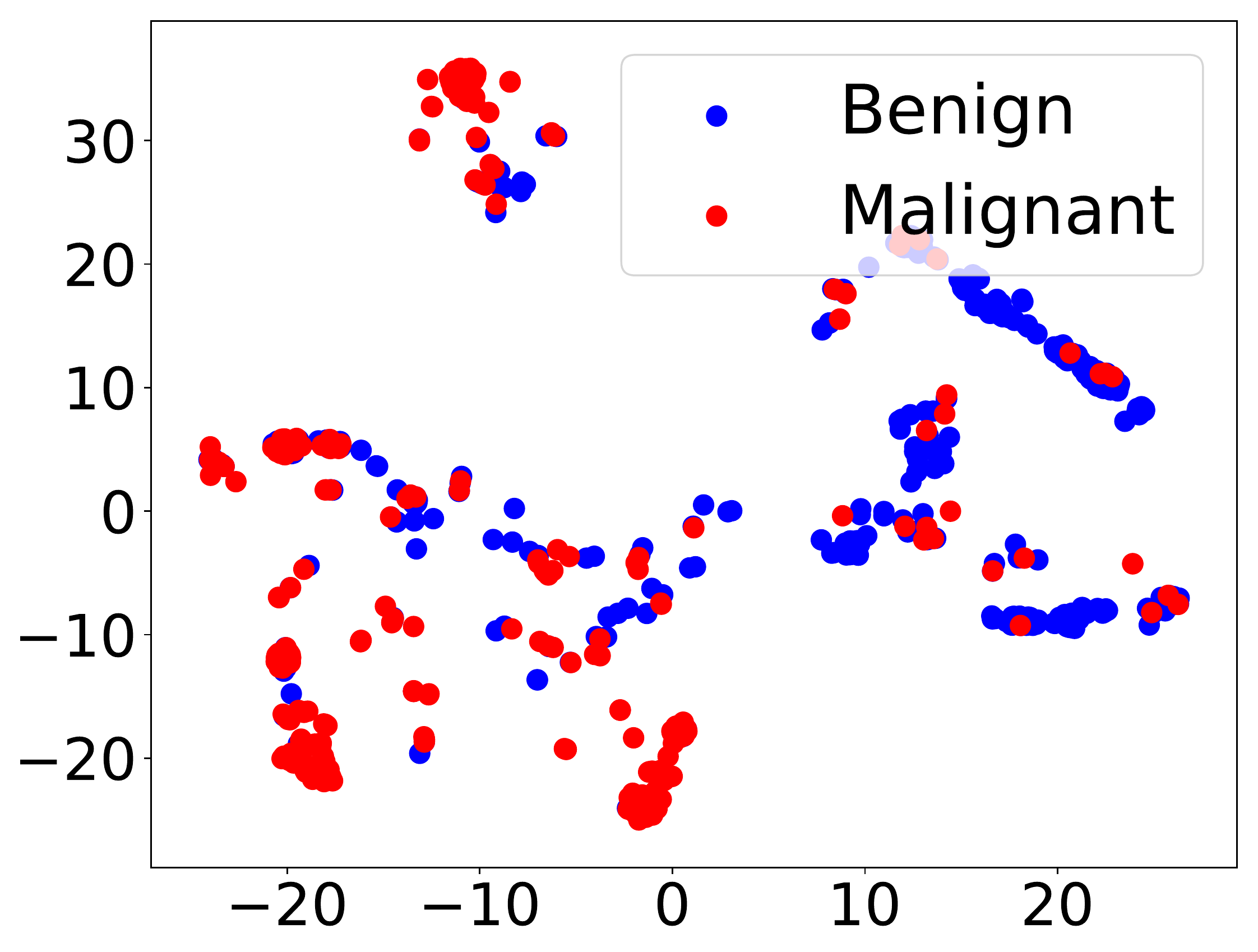}}
	
	\caption{$2D$ visualization of the embeddings, for two different datasets, obtained using the \textit{GrAMME-Fusion} architecture with parameters $T = 2$, $H = 1$ and $K = 5$ respectively. We also show the initial randomized features for reference.}
	\label{fig:embeddings}
\end{figure*}

\begin{table*}[t]
	\caption{Semi-Supervised learning performance of the proposed multi-layered attention architectures on the benchmark datasets. The results reported were obtained by averaging $20$ independent realizations.}
	\centering
	\label{table:results}
	\begin{tabular}{|c|c|c|c|c|c|c|c|}
		\hline
		& \multicolumn{5}{c|}{Baselines}                            &                                                                       &                                                                            \\ \cline{2-6}
		\multirow{-2}{*}{\begin{tabular}[c]{@{}c@{}}\% Nodes\\ (Train)\end{tabular}} & DeepWalk       & Node2Vec       & LINE  & PMNE(r) & MNE   & \multirow{-2}{*}{\begin{tabular}[c]{@{}c@{}}GrAMME\\ SG\end{tabular}} & \multirow{-2}{*}{\begin{tabular}[c]{@{}c@{}}GrAMME \\ Fusion\end{tabular}} \\ \hline \hline
		\multicolumn{8}{|l|}{\cellcolor[HTML]{EFEFEF}Vickers-Chan Dataset}                                                                                                                                                                                                                                 \\ \hline
		10\%                                                                         & 72             & 51.07          & 76.60 & 50.87   & 85.76 & 98.94                                                                 & \textbf{99.21}                                                             \\ \hline
		20\%                                                                         & 83.55          & 51.97          & 87.06 & 53.28   & 88.37 & 98.94                                                                 & \textbf{99.21}                                                             \\ \hline
		30\%                                                                         & 89.97          & 52.88          & 89.97 & 53.88   & 91.72 & 98.94                                                                 & \textbf{99.21}                                                             \\ \hline \hline
		\multicolumn{8}{|l|}{\cellcolor[HTML]{EFEFEF}Congress Votes Dataset}                                                                                                                                                                                                                          \\ \hline
		10\%                                                                         & 98.96          & 98.46          & 97.03 & 98.88   & 95.62 & \textbf{100}                                                          & \textbf{100}                                                               \\ \hline
		20\%                                                                         & 99.69          & 99.50          & 98.80 & 99.75   & 97.56 & \textbf{100}                                                          & \textbf{100}                                                               \\ \hline
		30\%                                                                         & 99.99          & 99.55          & 99.70 & 99.77   & 98.37 & \textbf{100}                                                          & \textbf{100}                                                               \\ \hline \hline
		\multicolumn{8}{|l|}{\cellcolor[HTML]{EFEFEF}Leskovec-Ng Dataset}                                                                                                                                                                                                                             \\ \hline
		10\%                                                                         & 91.16          & 81.76          & 68.54 & 85.58   & 73.34 & 91.56                                                                 & \textbf{93.32}                                                             \\ \hline
		20\%                                                                         & 96.35          & 85.41          & 77.39 & 89.71   & 85.79 & 96.25                                                                 & \textbf{97.62}                                                             \\ \hline
		30\%                                                                         & 98.31          & 86.99          & 83.58 & 91.35   & 89.90 & 98.30                                                                 & \textbf{98.73}                                                             \\ \hline \hline
		\multicolumn{8}{|l|}{\cellcolor[HTML]{EFEFEF}Reinnovation Dataset}                                                                                                                                                                                                                            \\ \hline
		10\%                                                                         & 72.02          & 72.18          & 51.98 & 70.76   & 72.51 & \textbf{76.42}                                                        & 75.28                                                                      \\ \hline
		20\%                                                                         & 73.13          & 74.04          & 55.21 & 73.45   & 75.40 & \textbf{80.72}                                                        & 79                                                                         \\ \hline
		30\%                                                                         & 76.02          & 76.13          & 60.13 & 75.29   & 74.72 & \textbf{83.16}                                                        & 80.95                                                                      \\ \hline \hline
		\multicolumn{8}{|l|}{\cellcolor[HTML]{EFEFEF}Mammography Dataset}                                                                                                                                                                                                                             \\ \hline
		10\%                                                                         & 75.72          & 76.38          & 76.39 & 76.48   & 75.13 & 82.27                                                                 & \textbf{82.63}                                                             \\ \hline
		20\%                                                                         & 73.99          & 77.41          & 75.40 & 76.91   & 76.72 & 83.01                                                                 & \textbf{83.28}                                                             \\ \hline
		30\%                                                                         & 74.13          & 77.82          & 76.16 & 75.51   & 77.59 & 83.06                                                                 & \textbf{83.75}                                                             \\ \hline \hline
		\multicolumn{8}{|l|}{\cellcolor[HTML]{EFEFEF}CKM (Social) Dataset}                                                                                                                                                                                                                            \\ \hline
		10\%                                                                         & 97.31          & 95.70          & 90.88 & 97.42   & 92.86 & 96.65                                                                 & \textbf{98.66}                                                             \\ \hline
		20\%                                                                         & 98.12          & 97.92          & 94.35 & 98.20   & 95.27 & \textbf{99.14}                                                        & 98.91                                                                      \\ \hline
		30\%                                                                         & 99.08          & 98.34          & 96.32 & 98.34   & 96.83 & 99.19                                                                 & \textbf{99.68}                                                             \\ \hline \hline
		\multicolumn{8}{|l|}{\cellcolor[HTML]{EFEFEF}Balance Scale Dataset}                                                                                                                                                                                                                           \\ \hline
		10\%                                                                         & 81.07 & 80.58          & 54.08 & \textbf{81.85}   & 77.71 & 77.67                                                                 & 80.15                                                                      \\ \hline
		20\%                                                                         & 86.15          & {86.22} & 58.95 & \textbf{88.74}   & 80.31 & 78.67                                                                 & {86.58}                                                                      \\ \hline
		30\%                                                                         & 87.27          & {88.61} & 64.44 & \textbf{89.87}   & 83.34 & 79.10                                                                 & {88.72}                                                                      \\ \hline 
	\end{tabular}
\end{table*}

\subsection{Experiment Setup}
In this section, we describe the experiment setup in detail for both the baseline methods and the proposed models. We run our experiments in a transductive learning setting. For each dataset, we choose a fixed amount of labeled nodes uniformly at random, while the remaining nodes are used for performance evaluation. In order to study the sensitivity of the proposed approaches over varying levels of labeled data availability, we varied the percentage of train nodes from $10\%$ to $30\%$. We repeated the experiments over $20$ independent realizations of train-test splits, and we report the average performance in all cases. The performance of the algorithms were measured using the overall accuracy score.

Since the first three baseline methods (DeepWalk, Node2Vec, LINE) are single-layer graph embedding techniques, we treat each layer in the multi-layered graph data as independent, and obtain embeddings for the layers separately. Subsequently, we average the embeddings for each node and build a logistic regression classifier to perform label prediction. For DeepWalk and Node2Vec, we set the embedding dimension to $128$, the window size to $10$ and the number of random walks to $80$. For LINE, we fixed the embedding dimension at $100$. Among the three variants of PMNE~\cite{liu2017principled}, namely network aggregation, Co-analysis, and result aggregation, we report the results only for the result aggregation method, as it often outperforms other variants. The hyper-parameter values for this method were chosen following the original paper. For MNE, a common embedding size of $200$ and a layer specific embedding size of $10$ were used.

For both of the proposed approaches, we considered architectures with $T=2$ attention layers, and fixed the input feature dimension $D = 64$. The number of hidden dimensions were fixed at $32$. For the GrAMME-SG architecture, we used $H = 2$ attention heads and a single fusion head. On the other hand, in GrAMME-Fusion, we set $H=2$ for each of the layers and in the supra fusion layer, we used $K = 5$ fusion heads. All networks were trained with the Adam optimizer, with the learning rate fixed at $0.001$.

\subsection{Results}
Table \ref{table:results} summarizes the performance of our approaches on the $7$ multi-layered graph datasets, along with the baseline results. Figure \ref{fig:training} illustrates the convergence characteristics of the proposed GrAMME-Fusion architecture under different training settings for the \textit{Mammography} dataset. As it can be observed, even with the complex graph structure (around $2$ million edges), the proposed solutions demonstrate good convergence characteristics. 

From the reported results, we make the following observations: In all the datasets, the proposed attention-based approaches consistently outperform the baseline techniques, providing highly robust models even when the training size was fixed at $10\%$. For example, with the \textit{Vickers-Chan} dataset, both our approaches produce an improvement of over $25\%$ when compared to a weaker baseline such as DeepWalk, and about $14\%$ improvement over the state-of-the-art MNE technique. Even with challenging datasets such as \textit{Reinnovation} and \textit{Mammography} datasets, the proposed approaches achieve improvements of $4\%-10\%$ over the baseline methods. This clearly demonstrates the effectiveness of our multi-layered graph embedding approaches in scenarios with heterogeneous relationships. Note that, \textit{Balance Scale} dataset is the only case where we found the PMNE baseline to be superior to the proposed approaches, however by a very small margin.

Finally, we visualize the multi-layered graph embeddings to qualitatively understand the behavior of the proposed approach. More specifically, we show the $2-$D t-SNE visualizations of the hidden representations for \textit{Congress Votes} and \textit{Mammography} datasets, obtained using GrAMME-Fusion. Figure \ref{fig:embeddings} shows that initial random features and the learned representations, wherein the effectiveness of the attention mechanism in revealing the class structure is clearly evident.

\subsection{Comparing GrAMME-SG and GrAMME-Fusion}
GrAMME-SG operates under the assumption that information is shared between all layers in a multi-layered graph, and use attention models to infer the actual dependencies. GrAMME-Fusion on the other hand, builds only layer-wise attention models, and introduces a \textit{supra fusion} layer that exploits the most relevant inter-layer dependencies using only fusion heads. Though GrAMME-Fusion outperforms GrAMME-SG in most of the datasets considered in our evaluation, we believe this is due to the fact that GrAMME-SG over-parametrizes inter-layer dependencies and can sometimes produce noisy edges. Consequently, in scenarios where strong dependencies exists between layers, GrAMME-SG will be more appropriate. For example, with the re-innovation dataset, different layers represents each country's performance in diverse sectors such as infrastructure, institutions, labor market etc. A country which has very good infrastructure and is financially stable can be expected to have a superior labor market and high-quality institutions. As our experiments results show, in that case, GrAMME-SG produces the best performance.

We now present brief analysis of the time complexity for the proposed methods. At their core, an attention layer that takes in a single-layered graph with $D$ dimensional attributes and produces $d$ dimensional embeddings incurs a computational complexity of $\mathcal{O}(NDd + Md)$. Here, the first term corresponds to the linear feed-forward layer, while the second term accounts for the attention computation. Note, in cases where the graph is densely connected, the second term can dominate the complexity. For GrAMME-SG, we explicitly construct a supra-graph consisting a total of $NL$ nodes and $NL^2 + \sum_{l=1}^{L} M^{(l)}$ edges. Here, the first term corresponds to virtual pillar edges introduced across layers, while the second terms is the sum of layer-specific edges. The computational complexity of an attention head in GrAMME-SG can hence be expressed as $\mathcal{O}(NLDd + NL^2d + d\sum_{l=1}^{L} M^{(l)})$. Consequently, in this case, the number of nodes $N$ plays a more dominant role, when compared to the single-layered case. The flexibility gained in modeling dependencies across layers comes at the price of increased computational complexity, since we need to deal with a much larger graph.

On the other hand, GrAMME-Fusion is computationally efficient, since it employs multiple fusion heads (supra fusion layer), while simplifying the layer-wise attention models. The complexity of an attention head in this case is given as $\mathcal{O}(NDd + \bar{M}d)$, where $\bar{M}$ indicates $max(M^{(l)})$. Note that the time complexity is similar to that of single-layered graph.  Interestingly, with the GrAMME-Fusion architecture, increasing the number of attention heads $H$ does not lead to significant performance improvements, demonstrating the effectiveness of supra fusion layers. Note that, an attention head is computationally expensive when compared to a fusion head in the supra-fusion layer. Consequently, restricting $H=1$ and increasing the number of fusion heads $K$ leads to a graceful increase in the overall complexity.

More importantly, when compared to classical network embedding techniques, this approach is scalable to large-scale graphs, both in terms of $N$ and $L$, since we do not have deal with explicit decomposition of Laplacian matrices. Finally, similar to existing attention models, both the proposed approaches incur $\mathcal{O}(1)$ sequential computations and hence can be entirely parallelized.

\section{Related Work}
\label{sec:related}
In this section, we briefly review prior work on deep feature learning for graph datasets, and multi-layered graph analysis. Note that, the proposed techniques are built on the graph attention networks recently proposed by \cite{velickovic2017graph}.

\subsection{Feature Learning with Graph-Structured Data}
Performing data-driven feature learning with graph-structured data has gained lot of interest, thanks to the recent advances in generalizing deep learning to non-Euclidean domains. The earliest application of neural networks to graph data can be seen in \cite{frasconi1998general, sperduti1997supervised}, wherein recursive models were utilized to model dependencies. More formal generalizations of recurrent neural networks to graph analysis were later proposed in \cite{gori2005new, scarselli2009graph}. Given the success of convolutional neural networks in feature learning from data defined on regular grids (e.g. images), the next generation of graph neural networks focused on performing graph convolutions efficiently. This implied that the feature learning was carried out to transform signals defined at nodes into meaningful latent representations, akin to filtering of signals~\cite{shuman2013emerging}. Since the spatial convolution operation cannot be directly defined on arbitrary graphs, a variety of spectral domain and neighborhood based techniques have been developed.

Spectral approaches, as the name suggests operate using the spectral representation of graph signals, defined using the eigenvectors of the graph Laplacian. For example, in \cite{bruna2013spectral}, convolutions are realized as multiplications in the graph Fourier domain, However, since the filters cannot be spatially localized on arbitrary graphs, this relies on explicit computation of the spectrum based on matrix inversion. Consequently, special families of spatially localized filters have been considered. Examples include the localization technique in \cite{henaff2015deep}, and Chebyshev polynomial expansion based localization in \cite{defferrard2016convolutional}. Building upon this idea, Kipf and Welling \cite{kipf2016semi} introduced graph convolutional neural networks (GCN) using localized first-order approximation of spectral graph convolutions, wherein the filters operate within an one-step neighborhood, thus making it scalable to even large networks. On the other hand, with non-spectral approaches, convolutions are defined directly on graphs and they are capable of working with different sized neighborhoods. For example, localized spatial filters with different weight matrices for varying node degrees are learned in \cite{duvenaud2015convolutional}. Whereas, in approaches such as \cite{niepert2016learning} neighborhood for each node is normalized to achieve a fixed size neighborhood. More recently, attention models, which are commonly used to model temporal dependencies in sequence modeling tasks, were generalized to model neighborhood structure in graphs. More specifically, graph attention networks \cite{velickovic2017graph} employ dot product based self attention mechanisms to perform feature learning in semi-supervised learning problems. While these methods have produced state-of-the-art results in the case of single-layer graphs, to the best of our knowledge, no generalization exists for multi-layered graphs, which is the focus of this paper. In particular, we build solutions for scenarios where no explicit node attributes are available.

\subsection{Multi-layered graph analysis}
Analysis and inferencing with multi-layered graphs is a challenging, yet crucial problem in data mining. With each layer characterizing a specific kind of relationships, the multi-layered graph is a comprehensive representation of relationships between nodes, which can be utilized to gain insights about complex datasets. Although the multi-layered representation is more comprehensive, a question that naturally arises is how to effectively fuse the information. Most existing work in the literature focuses on community detection, and an important class of approaches tackle this problem through joint factorization of the multiple graph adjacency matrices to infer  embeddings~ \cite{tang2009clustering, dong2012clustering}. In~\cite{gligorijevic2016fusion}, the symmetric non-negative matrix tri-factorization algorithm is utilized in order to factorize the adjacencies into non-negative matrices including a shared cluster indicator matrix. Other alternative approaches include subgraph pattern mining \cite{zeng2006coherent,boden2012mining} and information-theoretic optimization based on Minimum Description Length \cite{papalexakis2013more}. A comprehensive survey studying the algorithms and datasets on this topic can be found in \cite{kim2015community}. In ~\cite{wang2012multimodal}, optimization of multi-modal graph-based regularization is performed for image re-ranking. Wang \textit{et al.}~\cite{wang2017learning} proposed a model which generalizes conventional graph-based semi-supervised learning mthods to a hierarchical approach. A unified optimization framework is developed in \cite{li2018multi} to model within-layer connections and cross-layer connections simultaneously, to generate node embeddings for interdependent networks. Recently, Song and Thiagarajan~\cite{song2018} proposed to generalize the DeepWalk algorithm to the case of multi-layered graphs, through optimization with proxy clustering costs, and showed the resulting embeddings produce state-of-the-art results. In contrast to these approaches, we consider the problem of semi-supervised learning and develop novel feature learning techniques for the multi-layered case.

\section{Conclusions}
In this paper, we introduced two novel architectures, GrAMME-SG and GrAMME-Fusion, for semi-supervised node classification with multi-layered graph data. Our architectures utilize randomized node attributes, and effectively fuse information from both within-layer and across-layer connectivities, through the use of a weighted attention mechanism. While GrAMME-SG provides complete flexibility by allowing virtual edges between all layers, GrAMME-Fusion exploits inter-layer dependencies using fusion heads, operating on layer-wise hidden representations. Experimental results show that our models consistently outperform existing node embedding techniques. As part of the future work, the proposed solution can be naturally extended to the cases of multi-modal networks and interdependent networks. Furthermore, studying the effectiveness of simple and scalable attention models in other challenging graph inferencing tasks such as multi-layered link prediction and influential node selection remains an important open problem.

\section*{Acknowledgments}
This work was supported in part by the Arizona State University SenSIP center. This work was performed under the auspices of the U.S. Department of Energy by Lawrence Livermore National Laboratory under Contract DE-AC52-07NA27344. We also thank Prasanna Sattigeri for the useful discussions, and sharing data.

\bibliographystyle{IEEEbib}
\bibliography{main.bib}

%\printbibliography

\end{document}